# Tuned and GPU-accelerated parallel data mining from comparable corpora

Krzysztof Wołk, Krzysztof Marasek

Department of Multimedia
Polish-Japanese Academy of Information Technology, Koszykowa 86, Warsaw
kwolk@pja.edu.pl

**Abstract.** The multilingual nature of the world makes translation a crucial requirement today. Parallel dictionaries constructed by humans are a widely-available resource, but they are limited and do not provide enough coverage for good quality translation purposes, due to out-of-vocabulary words and neologisms. This motivates the use of statistical translation systems, which are unfortunately dependent on the quantity and quality of training data. Such has a very limited availability especially for some languages and very narrow text domains. Is this research we present our improvements to Yalign's mining methodology by reimplementing the comparison algorithm, introducing a tuning scripts and by improving performance using GPU computing acceleration. The experiments are conducted on various text domains and bi-data is extracted from the Wikipedia dumps.

**Keywords**: Machine translation, comparable corpora, Machine learning, NLP, Knowledge-free learning.

## 1 Introduction

The aim of this study is the preparation of parallel and comparable corpora. This work improves SMT quality through the processing and filtering of parallel corpora and through extraction of additional data from the resulting comparable corpora. To enrich the language resources of SMT systems, various adaptation and interpolation techniques will be applied to the prepared data. Evaluation of SMT systems was performed on random samples of parallel data using automated algorithms to evaluate the quality and potential usability of the SMT systems' output [1].

As far as experiments are concerned, the Moses Statistical Machine Translation Toolkit software, as well as related tools and unique implementations of processing scripts for the Polish language, are used. Moreover, the multi-threaded implementation of the GIZA++ tool is employed to train models on parallel data and to perform their symmetrization at the phrase level. The SMT system is tuned using the Minimum Error Rate Training (MERT) tool, which through parallel data specifies the optimum weights for the trained models, improving the resulting translations. The statistical language models from single-language data are trained and smoothed using the SRI Language Modeling toolkit (SRILM). In addition, data from outside the thematic domain is adapted. In the case of parallel models, Moore-Levis Filtering is used, while single-language models are linearly interpolated [1].

Lastly, the Yalign parallel data mining tool is enhanced. Its speed is increased by reimplementing it in a multi-threaded manner and by employing graphics processing unit (GPU) horsepower for its calculations. Yalign quality is improved by using the Needleman-Wunch algorithm for sequence comparison and by developing a tuning script that adjusts mining parameters to specific domain requirements [2].

Such SMT systems and data are urgently required for many purposes, including web, medical text, and international translation services, for example, the error-free, real-time translation of European Parliament Proceedings. Nonetheless, no progress can be made if not enough parallel data is available. Solution to this problem is presented in the scope of this research.

## 2   State of the art

Perhaps the most common approach is based on the retrieval of cross-lingual information. In the second approach, source documents must be translated using any machine translation system. The documents translated in that process are then compared with documents written in the target language, to find the most similar document pairs.

An interesting idea for mining parallel data from Wikipedia was described in [3]. The authors propose two separate approaches. The first idea is to use an online machine translation (MT) system to translate Dutch Wikipedia pages into English, and they try to compare original EN pages with translated ones. The idea, although interesting, seems computationally infeasible, and it presents a chicken-or-egg problem. Their second approach uses a dictionary generated from Wikipedia titles and hyperlinks shared between documents. Unfortunately, the second method was reported to return numerous, noisy sentence pairs.

Yasuda and Sumita [4] proposed a MT bootstrapping framework based on statistics that generate a sentence-aligned corpus. Sentence alignment is achieved using a bilingual lexicon that is automatically updated by the aligned sentences. Their solution uses a corpus that has already been aligned for initial training. They showed that 10% of Japanese Wikipedia sentences have an English equivalent.

Interwiki links were leveraged by the approach of Tyers and Pienaar in [5]. Based on Wikipedia link structure, a bilingual dictionary is extracted. In their work they measured the average mismatch between linked Wikipedia pages for different languages. They found that their precision is about 69-92%.

In [6] the authors attempt to advance the state of art in parallel data mining by modeling document-level alignment using the observation that parallel sentences can most likely be found in close proximity. They also use annotation available on Wikipedia and an automatically-induced lexicon model. The authors report recall and precision of 90% and 80%, respectively.

The author of [7] introduces an automatic alignment method for parallel text fragments that uses a textual entailment technique and a phrase-based SMT system. The author states that significant improvements in SMT quality were obtained (BLEU increased by 1.73) by using this aligned data.

The authors in [8] propose obtaining only title and some meta-information, such as publication date and time for each document, instead of its full contents to reduce the cost of building the comparable corpora. The cosine similarity of the titles' term frequency vectors were used to match titles and the contents of matched pairs.

In the present research, the Yalign tool is used. The solution was far from perfect, but after improvements that were made during this research, it supplied the SMT systems with bi-sentences of good quality in a reasonable amount of time.

## 3  Parallel data mining

For the experiments in data mining, the TED lectures domain prepared for IWSLT 2014[1] evaluation campaign by the FBK[2], was chosen. This domain is very wide and covers many unrelated subject areas [9]. Narrower domains were selected as well. The first corpus is composed of documents from the European Medicines Agency (EMEA) [10]. The second corpus was extracted from the proceedings of the European Parliament (EUP) by Philipp Koehn (University of Edinburgh) [11]. In addition, experiments on the Basic Travel Expression Corpus (BTEC), tourism-related sentences, were also conducted [12]. Lastly, a big corpus obtained from the OpenSubtitles.org web page was used as an example of human dialogs. Table 1 provides details on the number of unique words (WORDS) and their forms, as well as the number of bilingual sentence pairs (PAIRS).

**Table 1.** Corpora specification

| CORPORA | PL WORDS | EN WORDS | PAIRS |
|---|---|---|---|
| BTEC | 50,782 | 24,662 | 220,730 |
| TED | 218,426 | 104,117 | 151,288 |
| EMEA | 148,230 | 109,361 | 1,046,764 |
| EUP | 311,654 | 136,597 | 632,565 |
| OPEN | 1,236,088 | 749,300 | 33,570,553 |

## 4  Yalign and improvements

The Yalign tool was designed to automate the parallel text mining process by finding sentences that are close translation matches from comparable corpora. This presents opportunities for harvesting parallel corpora from sources like translated documents and the web. In addition, Yalign is not limited to a particular language pair. However, alignment models for two selected languages must first be created [2].

Unfortunately, the Yalign tool is not computationally feasible for large-scale parallel data mining. The standard implementation accepts plain text or web links, which need to be accepted, as input, and the classifier is loaded into memory for each pair alignment. In addition, the Yalign software is single-threaded. To improve performance, a solution that supplies the Yalign tool with articles from the database within one session, with no need to reload the classifier each time, was developed. The developed solution also facilitated multi-threading and decreased the mining time by a factor of 6.1 (using a 4-core, 8-thread i7 CPU). The alignment algorithm was also reimplemented (Needleman-Wunch is used instead of A* Search) for better accuracy and to leverage the power of GPUs for additional computing requirements. The tuning algorithm was also implemented. The two NW algorithms, with and without GPU optimization, are conceptually identical, but the second has an advantage in efficiency, depending on the hardware, up to max(n, m) times. However, the results of the A* algorithm, if the similarity calculation and the gap penalty are defined as in the NW algorithm, will be the same only if there is an additional constraint on paths: Paths cannot go upward or leftward in the M matrix. Yalign does not impose these additional conditions, so in some scenarios, repetitions of the same phrase may appear. In fact, every time the algorithm decides to move up or left, it is coming back into the second and first sequence, respectively.

---

[1] http://www.iwslt.org

[2] http://www.fbk.eu

The quality of alignments in Yalign is defined by a tradeoff between precision and recall. The Yalign has two configurable variables:
- Threshold: the confidence threshold to accept an alignment as "good." A lower value means more precision and less recall. The "confidence" is a probability estimated from a support vector machine classifying "is a translation" or "is not a translation."
- Penalty: controls the amount of "skipping ahead" allowed in the alignment [2].

Both of these parameters are selected automatically during training, but they can be adjusted if necessary. The solution implemented in this research also introduces a tuning algorithm for those parameters, which allows better adjustment of them.

## 5  Evaluation of obtained comparable corpora

To evaluate the corpora, each corpus was divided into 200 segments, and 10 sentences were randomly selected from each segment. This methodology ensured that the test sets covered the entire corpus. The selected sentences were removed from the corpora. The testing system was trained with the baseline settings. In addition, a system was trained with extended data from the Wikipedia corpora. Lastly, Modified Moore-Levis Filtering was used for the Wikipedia corpora domain adaptation. The monolingual part of the corpora was used as language model and was adapted for each corpus by using linear interpolation [13].

The evaluation was conducted using test sets built from 2,000 randomly selected bi-sentences taken from each domain. For scoring purposes, four well-known metrics that show high correlation with human judgments were used. Among the commonly used SMT metrics are: Bilingual Evaluation Understudy (BLEU), the U.S. National Institute of Standards & Technology (NIST) metric, the Metric for Evaluation of Translation with Explicit Ordering (METEOR) and Translation Error Rate (TER) [13].

First, speed improvements were made by introducing multi-threading to the algorithm, using a database instead of plain text files or Internet links, and using GPU acceleration in sequence comparison. More importantly, two improvements were made to the quality and quantity of the mined data. The A* search algorithm [14] was modified to use Needleman-Wunch [15], and a tuning script of mining parameters was developed. During this empirical research, it was realized that Yalign suffers from a problem that produces different results and quality measures, depending on whether the system was trained from a foreign to a native language or vice versa. To cover as much parallel data as possible during the mining, it is necessary to train the classifiers bidirectionally for the language pairs of interest. By doing so, additional bi-sentences can be found. The data mining approaches used were: directional (PL->EN classifier) mining (MONO), bi-directional (additional EN->PL classifier) mining (BI), bi-directional mining with Yalign using a GPU-accelerated version of the Needleman-Wunch [16] algorithm (NW), and mining using a NW version of Yalign that was tuned (NWT). The results of such mining are shown in Table 2.

**Table 2.** Number of obtained Bi-Sentences

| Mining Method | Number of Bi-Sentences | Uniq PL Tokens | Uniq EN Tokens |
|---|---|---|---|
| MONO | 510,128 | 362071 | 361039 |
| BI | 530,480 | 380771 | 380008 |
| NW | 1,729,061 | 595064 | 574542 |
| NWT | 2,984,880 | 794478 | 764542 |

As presented in Table 2, each of the improvements increased the number of parallel sentences discovered. However, there is no indication of the quality of the obtained data, SMT improvements, or information regarding the computation time of the NW version of Yalign. To address these aspects, two additional experiments were conducted. First, in Table 3 a speed comparison is made using different versions of the Yalign Tool. A total of 1,000 comparable articles were randomly selected from Wikipedia and aligned using the native Yalign implementation (Yalign), multi-threaded implementation (M Yalign), Yalign with the Needleman-Wunch algorithm (NW Yalign), and Yalign with a GPU-accelerated Needleman-Wunch algorithm (GNW Yalign). Second, MT experiments were conducted to verify potential gains in translation quality on the data that was tuned and aligned using a different heuristic. The TED, EUP, EMEA and OPEN domains were used for this purpose. For each of the domains, the system was trained using baseline settings. The additional corpora were used in the experiments by adding parallel data to the training set using Modified Moore-Levis Filtering and by adding a monolingual language model with linear interpolation. The results are shown in Table 4, where BASE represents the baseline system; MONO, the system enhanced with a mono-directional classifier; BI, a system with bi-directional mining; NW, a system mined bi-directionally using the Needleman-Wunch algorithm; and TNW, a system with additionally tuned parameters.

**Table 3.** Computation Time of Different Yalign Version

| Mining Method | Computation Time [s] |
|---|---|
| YALIGN | 89,67 |
| M YALIGN | 14,7 |
| NW YALIGN | 17,3 |
| GNW YALIGN | 15,2 |

**Table 4.** Results of SMT Enhanced Comparable Corpora for PL to EN and PL to EN translation

|  |  | *PL to EN* | | | | *EN to PL* | | | |
|---|---|---|---|---|---|---|---|---|---|
|  |  | BLEU | NIST | TER | METEOR | BLEU | NIST | TER | METEOR |
| TED | BASE | 16,96 | 5,26 | 67,10 | 49,42 | 10,99 | 3,95 | 74,87 | 33,64 |
|  | MONO | 16,97 | 5,39 | 65,83 | 50,42 | 11,24 | 4,06 | 73,97 | 34,28 |
|  | BI | 17,34 | 5,37 | 66,57 | 50,54 | 11,54 | 4,02 | 73,75 | 34,43 |
|  | NW | 17,45 | 5,37 | 65,36 | 50,56 | 11,59 | 3,98 | 74,49 | 33,97 |
|  | TNW | 17,50 | 5,41 | 64,36 | 90,62 | 11,98 | 4,05 | 73,65 | 34,56 |
| EUP | BASE | 36,73 | 8,38 | 47,10 | 70,94 | 25,74 | 6,56 | 58,08 | 48,46 |
|  | MONO | 36,89 | 8,34 | 47,12 | 70,81 | 24,71 | 6,37 | 59,41 | 47,45 |
|  | BI | 36,56 | 8,34 | 47,33 | 70,56 | 24,63 | 6,35 | 59,73 | 46,98 |
|  | NW | 35,69 | 8,22 | 48,26 | 69,92 | 24,13 | 6,29 | 60,23 | 46,78 |
|  | TNW | 35,26 | 8,15 | 48,76 | 69,58 | 24,32 | 6,33 | 60,01 | 47,17 |
| EMEA | BASE | 62,60 | 10,19 | 36,06 | 77,48 | 56,39 | 9,41 | 40,88 | 70,38 |
|  | MONO | 62,48 | 10,20 | 36,29 | 77,48 | 55,38 | 9,25 | 42,37 | 69,35 |
|  | BI | 62,62 | 10,22 | 35,89 | 77,61 | 56,09 | 9,32 | 41,62 | 69,89 |
|  | NW | 62,69 | 10,27 | 35,49 | 77,85 | 55,80 | 9,30 | 42,10 | 69,54 |
|  | TNW | 62,88 | 10,26 | 35,42 | 77,96 | 55,63 | 9,31 | 41,91 | 69,65 |
| OPEN | BASE | 64,58 | 9,47 | 33,74 | 76,71 | 31,55 | 5,46 | 62,24 | 47,47 |
|  | MONO | 65,77 | 9,72 | 32,86 | 77,14 | 31,27 | 5,45 | 62,43 | 47,28 |
|  | BI | 65,87 | 9,71 | 33,11 | 76,88 | 31,23 | 5,40 | 62,70 | 47,03 |
|  | NW | 65,79 | 9,73 | 33,07 | 77,31 | 31,47 | 5,46 | 62,32 | 47,39 |
|  | TNW | 65,91 | 9,78 | 32,22 | 77,36 | 31,80 | 5,48 | 62,27 | 47,47 |

The results indicate that multi-threading significantly improved speed, which is very important for large-scale mining. As anticipated, the Needleman-Wunch algorithm decreases speed (that is why authors of the Yalign did not use it, in the first place). However, GPU acceleration makes it possible to obtain performance almost as fast as that of the multi-threaded A* version. It must be noted that the mining time may significantly differ when the alignment matrix is big (text is long). The experiments were conducted on a hyper-threaded Intel Core i7 CPU and a GeForce GTX 660 GPU. The quality of the data obtained with the NW algorithm version, as well as the TNW version, seems promising. Slight improvements in the translation quality were observed, but more importantly, much more parallel data was obtained.

It was decided to train an SMT system using only data extracted from comparable corpora (not using the original in domain data) to verify the results. The mined data was used also as a language model. The evaluation was conducted using the same test sets shown in Table 4. The results are presented in Table 5, where BASE indicates the results for the baseline system trained on the original in-domain data; MONO, a system trained only on mined data in one direction; BI, a system trained on data mined in two directions with duplicate segments removed; NW, a system using bi-directionally mined data with the Needleman-Wunch algorithm; and TNW, a system with additionally tuned parameters.

**Table 5.** SMT Results Using only Comparable Corpora for PL to EN translation

|  |  | *PL to EN* | | | | *EN to PL* | | | |
|---|---|---|---|---|---|---|---|---|---|
|  |  | BLEU | NIST | TER | METEOR | BLEU | NIST | TER | METEOR |
| TED | BASE | 16,96 | 5,26 | 67,10 | 49,42 | 10,99 | 3,95 | 74,87 | 33,64 |
|      | MONO | 12,91 | 4,57  | 71,50 | 44,01 | 7,89  | 3,22 | 83,90 | 29,20 |
|      | BI   | 12,90 | 4,58  | 71,13 | 43,99 | 7,98  | 3,27 | 84,22 | 29,09 |
|      | NW   | 13,28 | 4,62  | 71,96 | 44,47 | 8,50  | 3,28 | 83,02 | 29,88 |
|      | TNW  | 13,94 | 4,68  | 71,50 | 45,07 | 9,15  | 3,38 | 78,75 | 30,08 |
| EUP  | BASE | 36,73 | 8,38  | 47,10 | 70,94 | 25,74 | 6,56 | 58,08 | 48,46 |
|      | MONO | 21,82 | 6,09  | 62,85 | 56,40 | 14,48 | 4,76 | 70,64 | 36,62 |
|      | BI   | 21,24 | 6,03  | 63,27 | 55,88 | 13,91 | 4,67 | 71,32 | 35,83 |
|      | NW   | 20,20 | 5,88  | 64,24 | 54,38 | 13,13 | 4,54 | 72,14 | 35,03 |
|      | TNW  | 20,42 | 5,88  | 63,95 | 54,65 | 13,41 | 4,58 | 71,83 | 35,34 |
| EMEA | BASE | 62,60 | 10,19 | 36,06 | 77,48 | 56,39 | 9,41 | 40,88 | 70,38 |
|      | MONO | 21,71 | 5,09  | 74,30 | 44,22 | 19,11 | 4,73 | 74,77 | 37,34 |
|      | BI   | 21,45 | 5,06  | 73,74 | 44,01 | 18,65 | 4,60 | 75,16 | 36,91 |
|      | NW   | 21,47 | 5,06  | 73,81 | 44,14 | 18,60 | 4,53 | 76,19 | 36,30 |
|      | TNW  | 22,64 | 5,30  | 72,98 | 45,52 | 18,58 | 4,48 | 76,60 | 36,28 |
| OPEN | BASE | 64,58 | 9,47  | 33,74 | 76,71 | 31,55 | 5,46 | 62,24 | 47,47 |
|      | MONO | 11,53 | 3,34  | 78,06 | 34,71 | 7,95  | 2,40 | 88,55 | 24,37 |
|      | BI   | 11,64 | 3,25  | 82,38 | 33,88 | 8,20  | 2,40 | 89,51 | 24,49 |
|      | NW   | 11,64 | 3,32  | 81,48 | 34,62 | 9,02  | 2,52 | 86,14 | 25,01 |

## 7 Conclusions

The results for SMT systems based only on mined data are not very surprising. First, they confirm the quality and high level of parallelism of the corpora. This can be concluded from the translation quality, especially for the TED data set. Only two BLEU scoring anomalies were observed when comparing systems strictly trained on in-domain (TED) data and mined data for EN to PL translation. It also seems reasonable that the best SMT scores were obtained on TED data. This data set is the most similar to the Wikipedia articles, overlapping with it on many topics. In addition, the Yalign classifier trained on the TED data set recognized most of the parallel sentences. The results show that the METEOR metric, in some cases, increases when the other metrics decrease. The most likely explanation for this is that other metrics suffer, in comparison to METEOR, from the lack of a scoring mechanism for synonyms. Wikipedia is a very wide domain, not only in terms of its topics, but also its vocabulary. This leads to a conclusion that mined corpora is good source for extending sparse text domains. It is also the reason why test sets originating from wide domains outscore those of narrow domains and also why training on a larger mined data set sometimes slightly decreases the results from very specific domains. Nonetheless, in many cases after manual analysis was conducted, the translations were good, but the automatic metrics were lower due to the usage of synonyms.

In addition, it was proven that mining data using two classifiers trained from a foreign to a native language and vice versa can significantly improve data quantity, even though some repetition is possible. Such bi-directional mining, which is logical, found additional data mostly for wide domains. In narrow text domains, the potential gain is small. From a practical point of view, the method requires neither expensive training nor language-specific grammatical resources, but it produces satisfying results. It is possible to replicate such mining for any language pair or text domain, or for any reasonably comparable input data.

The results presented in Table 5 show a slight improvement in translation quality, which verifies that the improvements to Yalign positively impact the overall mining process. It must be noted that the above mining experiments were conducted using a classifier trained only on the TED data. This is why the improvements are very visible on this corpora and less visible on other corpora. What is more improvements obtained by enriching training set were observed mostly on wide-domain text, while for narrow domains the effects were negative. The improvements were mostly observed on the Ted data set, because the classifier was only trained on text samples. Nonetheless, regardless the text domain tuning algorithm proved to always improve the transaction quality.